\documentclass[10pt,twocolumn,letterpaper]{article}

\usepackage{cvpr}              

\usepackage{array}
\usepackage{bm}

\newcommand{\figlabel}{Fig.\xspace}

\newcommand{\inlinesection}[1]{\noindent\textbf{#1.}}

\newcommand{\supp}{\textit{Appendix}\xspace}

\definecolor{Highlight}{HTML}{39b54a}  

\usepackage{amssymb}
\usepackage{pifont}
\usepackage{algorithm}
\usepackage{listings}

\newif\ifdraft
\drafttrue
\ifdraft
\newcommand{\gqc}[1]{{\color{orange}[\textbf{Gordon:} #1]}}
\newcommand{\kac}[1]{{\color{purple}[\textbf{Kfir:} #1]}}
\newcommand{\dcc}[1]{{\color{red}[\textbf{Danny:} #1]}}
\newcommand{\jacksoncomment}[1]{{\color{blue}[\textbf{Jackson:} #1]}}
\newcommand{\opc}[1]{{\color{teal}[\textbf{Or:} #1]}}
\newcommand{\dosc}[1]{{\color{magenta}[\textbf{Daniil:} #1]}}

\newcommand{\todo}[1]{{\textbf{\color{red}[TODO: #1]}}}

\else
\newcommand{\gqc}[1]{}
\newcommand{\kac}[1]{}
\newcommand{\dcc}[1]{}
\newcommand{\jacksoncomment}[1]{}
\newcommand{\opc}[1]{}
\newcommand{\todo}[1]{}

\newcommand{\dosc}[1]{}
\fi
\usepackage{etoolbox}
\makeatletter
\AfterEndEnvironment{algorithm}{\let\@algcomment\relax}
\AtEndEnvironment{algorithm}{\kern2pt\hrule\relax\vskip3pt\@algcomment}
\let\@algcomment\relax
\newcommand\algcomment[1]{\def\@algcomment{\footnotesize#1}}
\renewcommand\fs@ruled{\def\@fs@cfont{\bfseries}\let\@fs@capt\floatc@ruled
  \def\@fs@pre{\hrule height.8pt depth0pt \kern2pt}%
  \def\@fs@post{}%
  \def\@fs@mid{\kern2pt\hrule\kern2pt}%
  \let\@fs@iftopcapt\iftrue}
\makeatother
\newcommand{\cmmnt}[1]{}

\usepackage[normalem]{ulem}

\usepackage{gensymb}
\usepackage{textcomp}

\definecolor{cvprblue}{rgb}{0.21,0.49,0.74}
\usepackage[pagebackref,breaklinks,colorlinks,
            citecolor=cvprblue,    
           ]{hyperref}
\usepackage{arydshln}

\def\paperID{5789} 
\def\confName{CVPR}
\def\confYear{2025}

\newcommand{\webpage}{\url{https://snap-research.github.io/Omni-ID/}~}

\newcommand{\methodname}{Omni-ID}
\newcommand{\methodnametwo}{Omni-ID} 

\begin{document}

\title{
Omni-ID: Holistic Identity Representation Designed for Generative Tasks\\
}

\author{
Guocheng Qian \quad
Kuan-Chieh Wang \quad
Or Patashnik \quad
Negin Heravi \\
Daniil Ostashev \quad
Sergey Tulyakov \quad
Daniel Cohen-Or \quad
Kfir Aberman \\
Snap Research 
\\
{\tt\small \webpage}
}

\maketitle

\begin{abstract}
We introduce Omni-ID, a novel facial representation designed specifically for generative tasks. Omni-ID encodes holistic information about an individual's appearance across diverse expressions and poses within a fixed-size representation. It consolidates information from a varied number of unstructured input images into a structured representation, where each entry represents certain global or local identity features. Our approach uses a few-to-many identity reconstruction training paradigm, where a limited set of input images is used to reconstruct multiple target images of the same individual in various poses and expressions. A multi-decoder framework is introduced to leverage the complementary strengths of diverse decoders during training. Unlike conventional representations, such as ArcFace and CLIP, which are typically learned through discriminative or contrastive objectives, Omni-ID is optimized with a generative objective, resulting in a more comprehensive and nuanced identity capture for generative tasks. Trained on our MFHQ dataset -- a multi-view facial image collection, Omni-ID demonstrates substantial improvements over conventional representations across various generative tasks.
\end{abstract}

\begin{figure}
    \centering
    \includegraphics[width=\linewidth]{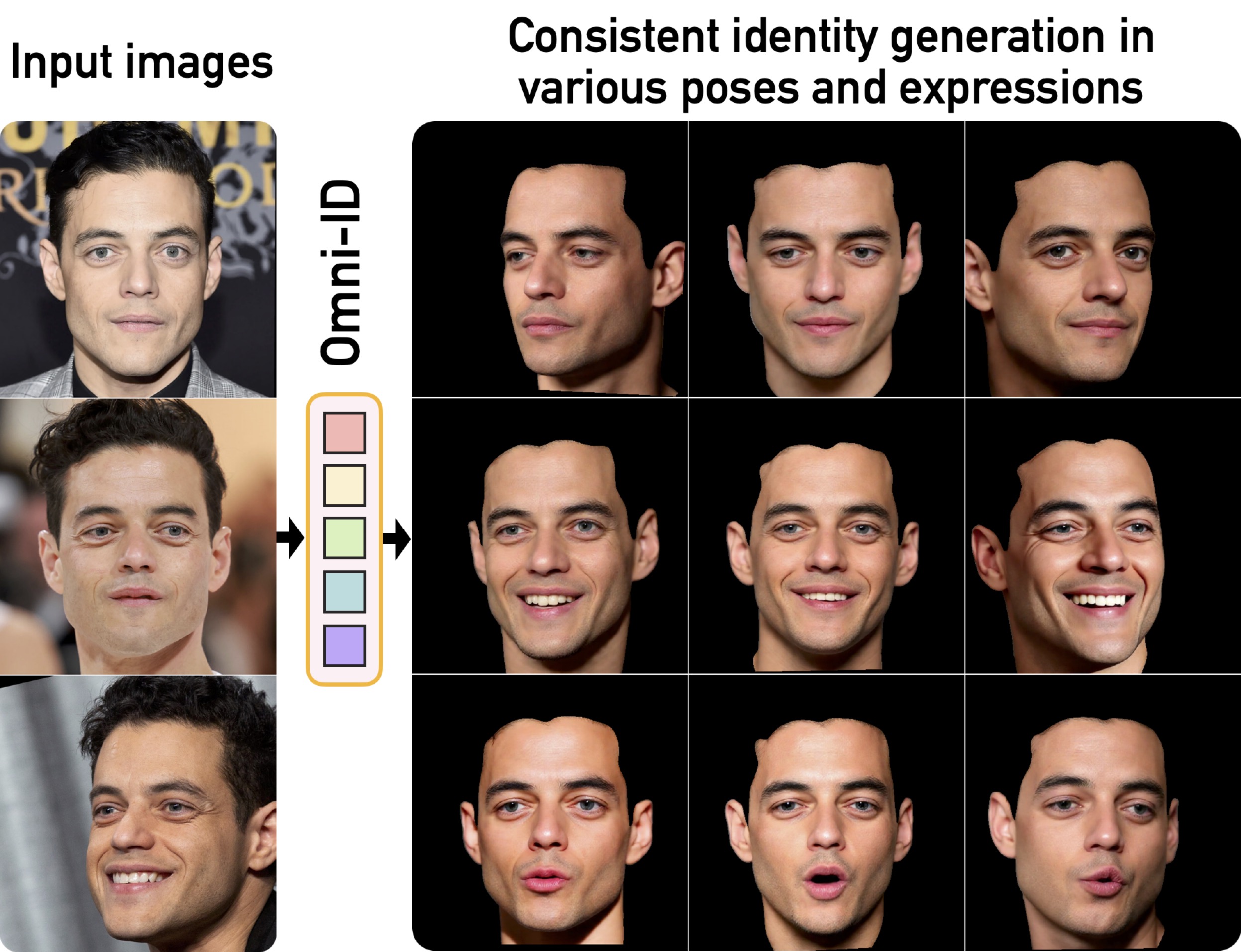}
    \caption{
    \textbf{Omni-ID} is a facial representation that consolidates information from a varied number of images of an individual into a fixed-size, structured encoding. Each element of this encoding captures specific global or local identity features, enabling high-fidelity generation in new poses, expressions, and capturing identity-consistent variations. 
    }
    \label{fig:teaser}
\end{figure}

\section{Introduction}\label{sec:intro}

Generating images that faithfully represent an individual's identity requires a face encoding capable of depicting nuanced details across diverse poses and facial expressions.
However, a significant limitation of existing facial representations \cite{CosFace, ArcFace, CLIP, FaRL} is their reliance on single-image encodings, which fundamentally lack holistic information about one's appearance. 
For example, an image of someone in a frontal pose with a neutral expression reveals little about how they look when smiling, frowning, or viewed from their profile.

\begin{figure}
    \centering
    \includegraphics[width=\linewidth]{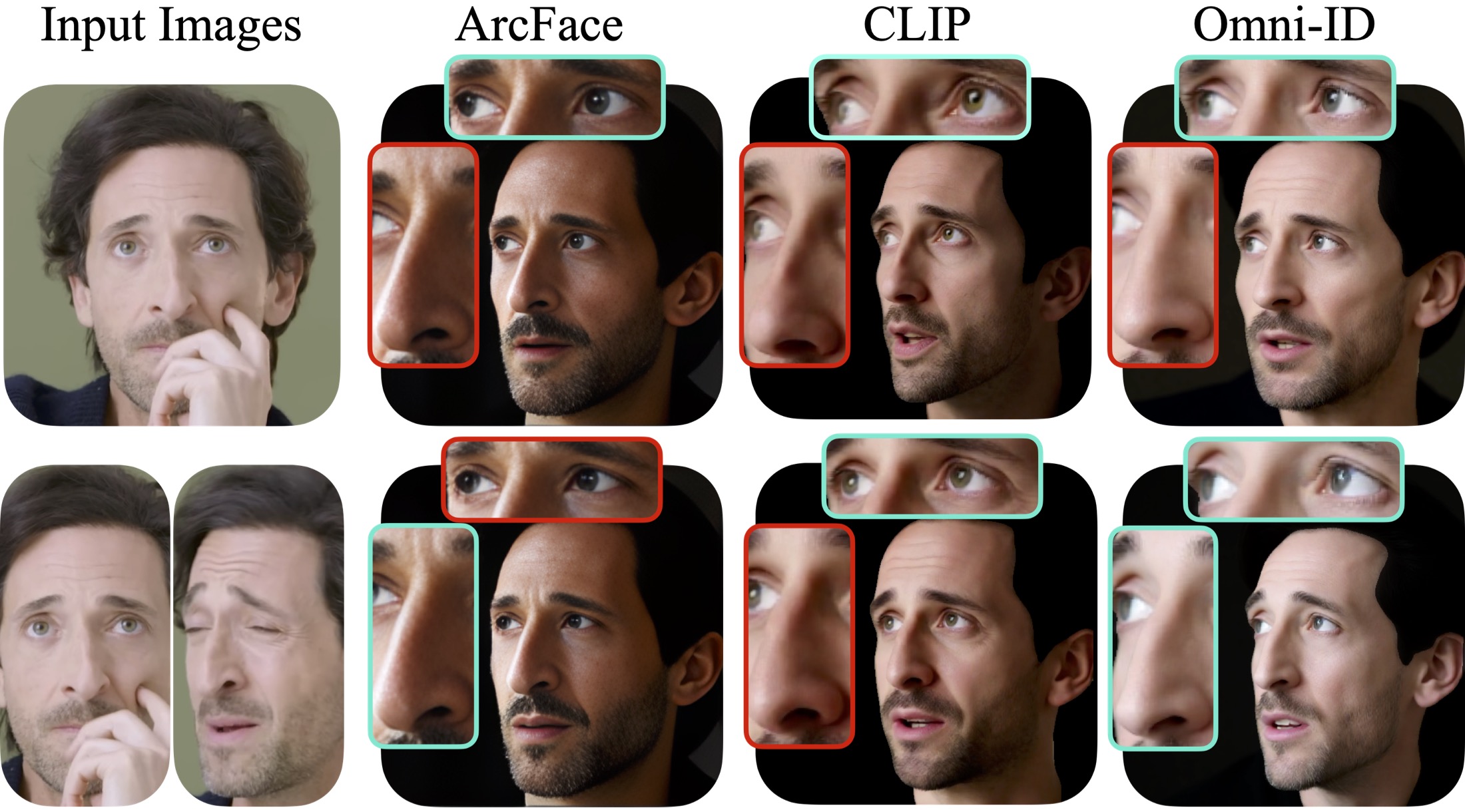}
    \caption{\textbf{Face generation comparison of different facial representations} with single input (top row) and two inputs (bottom row). We evaluate different facial representations by training an IP-adapter~\cite{ye2023ip-adapter} on FLUX~\cite{FLUX} with each representation. 
    It can be seen that single-instance representations such as ArcFace and CLIP struggle to combine unique features appear in each observation (e.g., eye color and nose shape), whereas our Omni-ID, designed with a \emph{few-to-many} generative objective, improves identity representation with each additional view, unifying unique attribute from multiple views into a single representation. 
    }
    \label{fig:motivation}
\end{figure}

Furthermore, existing face representation methods, typically derived from networks optimized for discriminative tasks such as ArcFace~\cite{ArcFace} or text-aligned image encoders like CLIP~\cite{CLIP}, are not well suited for generative applications. 
Intuitively, Information Bottleneck Principle \cite{achille2017emergence} suggests that subtle variations critical for generation but irrelevant to class boundary are likely lost in discriminative training. 
Consequently, discriminative features struggled to capture the subtle nuances that define a person's unique identity especially across different poses and expressions. Refer to \cref{fig:motivation,fig:sota-person,fig:sota-face} for examples of an inaccurate nose, missing beards, and an inaccurate head shape when using discriminative features for generation.

In this work, we introduce \emph{Omni-ID}, a facial identity representation designed for generative tasks. Omni-ID encodes a varied number of images of an individual into a compact, fixed-size representation, capturing the individual in diverse expressions and poses as shown in \cref{fig:teaser}. This enriched encoding can be used in a wide range of generative tasks, maximizing the faithful preservation of the individual's subtle details across various contexts.

\begin{figure}[t] 
\centering
\includegraphics[width=1.0\linewidth]{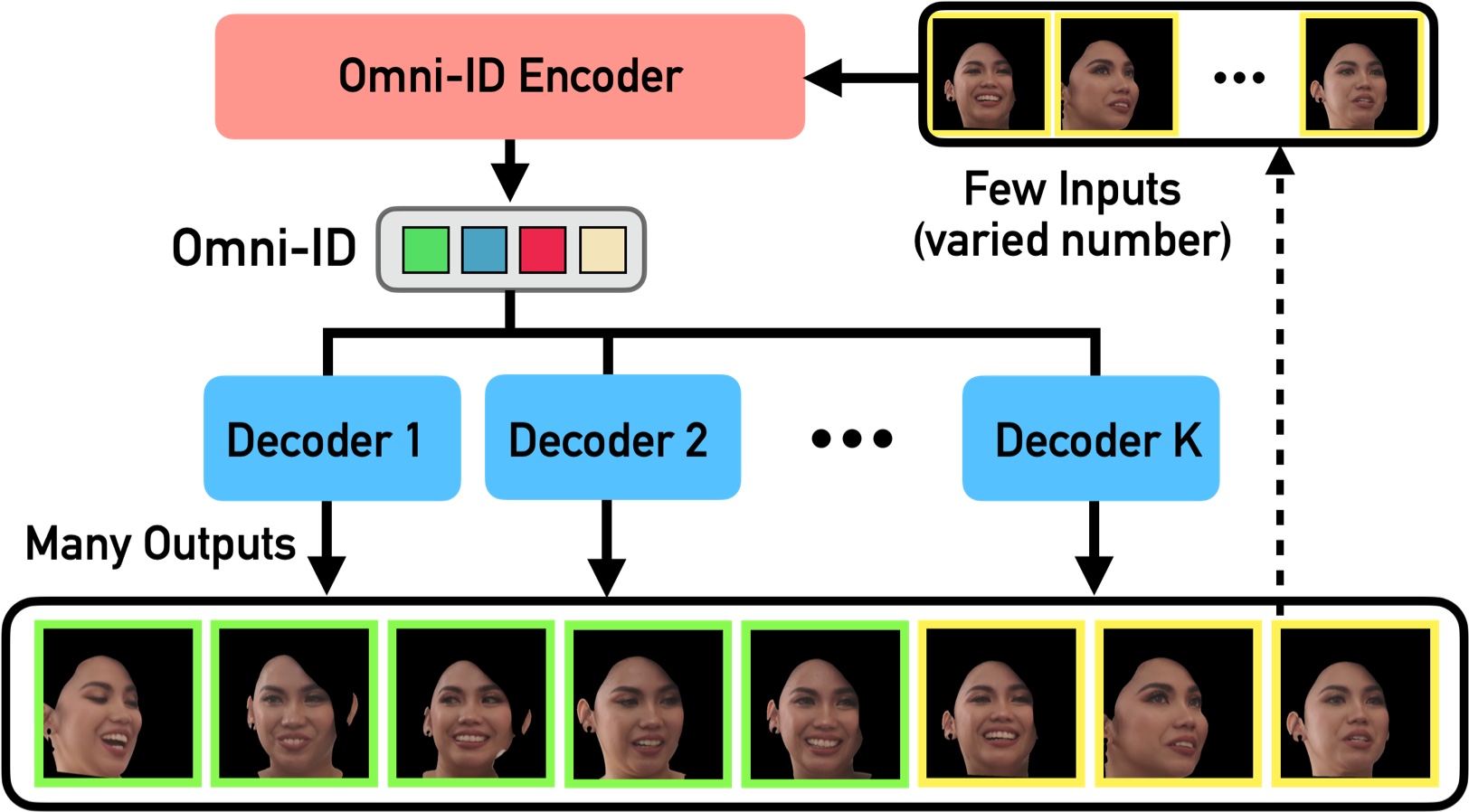}
\caption{
\methodname~employs a \textit{\textbf{multi-decoder few-to-many identity reconstruction}} training strategy, incorporating three key design features: (1) An encoder that learns a unified, fixed-size identity representation  from a varied number of inputs; (2) A few-to-many identity reconstruction task, designed to generate multiple faces of an individual in various poses and expressions from a limited set of samples of the same individual; (3) A multi-decoder training strategy that combines the unique strengths of various decoders while mitigating the limitations of any single decoder.
}
\label{fig:overall_pipeline}
\end{figure}

At the core of the approach lies our Omni-ID encoder trained within a generative framework (illustrated in \cref{fig:overall_pipeline}) that leverages two key ideas.
First, the framework uses a \emph{few-to-many identity reconstruction} training paradigm that not only reconstructs the input images but also a diverse range of other images of the same identity in various contexts, poses, and expressions. This strategy encourages the representation to capture essential identity features observed across different conditions while mitigating overfitting to specific attributes of any single input image.
Second, our framework employs a \emph{multi-decoder} training objective that combines the unique strengths of various decoders, such as improved fidelity or reduced identity leakage, while mitigating the limitations of any single decoder. This enables leveraging the detailed facial information present in the input images to the greatest feasible degree and results in a more robust encoding that effectively generalizes across various generative applications.

Our Omni-ID encoder is transformer-based and employs a fixed-size set of learnable queries to produce a consistent, fixed-size representation of an individual's identity. 
The fixed-size representation is essential for downstream applications, as it establishes a `\textit{structured}' encoding where each feature within the representation can correspond to specific identity attributes, such as different facial regions as visualized in \cref{fig:attention}. Structured representations allow downstream tasks to focus on
learning from the distilled identity features without being distracted by the noise and variability present in individual input images.

To validate Omni-ID’s effectiveness, we conduct extensive experiments comparing our method against state-of-the-art baselines, including ArcFace~\cite{ArcFace} and CLIP~\cite{CLIP} representations. Notably, with our representation the generation quality scales with the number of input images, enabling it to capture a more comprehensive view of the individual, as shown in \cref{fig:motivation}. In addition, we demonstrate Omni-ID's superiority in two widely used face generative tasks: controllable face synthesis and personalized text-to-image generation. Extensive experiments show that Omni-ID significantly improves identity preservation across a range of poses, expressions, and contexts, achieving higher fidelity in generating photorealistic images.

\section{Related work}\label{sec:related}

\noindent\textbf{Face representation} provides a foundation for 3D face reconstruction, accurately distinguishing identities and synthesizing realistic face images.  Parametric 3D Morphable Models \cite{Blanz3DMM, FaceWarehouse, FLAME:SiggraphAsia2017} have been historically used to represent face shape geometry through identity and expression blendshapes combined with pose. However, these representations are coarse and lack the appearance details needed for photo-realistic generation \cite{feng2021deca, danvevcek2022emoca, Retsinas24Smirk}. In recognition tasks, approaches like CosFace \cite{CosFace} and ArcFace \cite{ArcFace} have improved identity discrimination by utilizing a margin loss, which enhances intra-class compactness and inter-class separability, and have also been widely applied in generation tasks. 
More recently, FaRL \cite{FaRL} creates a more descriptive facial token representations by fine-tuning the pretrained visual model, CLIP \cite{CLIP} on large-scale face-text paired datasets.
In contrast to these discriminative or contrastive  facial identity representations, our Omni-ID is optimized with a generative objective, resulting in a more nuanced identity representation well suited for generative tasks.

\noindent\textbf{Face synthesis} has evolved significantly, starting with StyleGAN \cite{karras2019style}, which set new standards for high-quality, realistic face generation through a well-structured latent space.  However, StyleGAN offered limited control over individual facial features. To address this, researchers introduced methods to invert real images into StyleGAN’s latent space \cite{abdal2019image2stylegan, harkonen2020ganspace, abdal2020styleflow,tov2021designing,richardson2021encoding}, enabling attribute manipulation by altering latent codes. 
Given recent advances in generative models, diffusion~\cite{ho2020denoising, Song2020DenoisingDI, Dhariwal2021DiffusionMB, Ho2022ClassifierFreeDG,nitzan2022mystyle} and flow-based~\cite{Rombach2021HighResolutionIS} generative models offer even higher quality generations than GAN-based approaches.  In~\cref{sec:exp:posedface} we study controllable face generation and show improved identity fidelity when using our proposed face representation.

\noindent\textbf{Personalized text-to-image generation} embeds specific visual elements or concepts that are unique to individual users or classes of images to text-to-image models ~\cite{ho2020denoising,ldm}. Early works focused on introducing new tokens or fine-tuning model weights to represent personalized content while maintaining the prior of the model~\cite{ruiz2023dreambooth,gal2022image,voynov2023p+, hu2022lora}.
Since then, feedforward methods were introduced to reduce the computational cost from per-subject optimization \cite{ruiz2023hyperdreambooth, ye2023ip-adapter}. These techniques typically utilized encoders~\cite{gal2023encoder} or adapters~\cite{shi2023instantbooth,wang2024moa} that process images into representations then directly inject into text-to-image models during inference. 
A major focus of research has been placed on personalizing text-to-image models for human faces. 
IP-Adapter \cite{ye2023ip-adapter} proposed to inject faces through decoupled attention layers. Follow-ups improve identity preservation and controls by ControlNet \cite{ControlNet, wang2024instantid}, text embedding merging \cite{li2023photomaker}, and face identity loss \cite{gal2024lcm, PuLID}.  
However, existing personalization approaches rely on facial representations derived from single-instance encodings, extracted from networks trained with discriminative objectives. Orthogonal to these efforts, our work focuses on identity representation compatible with the different approaches.  We show that Omni-ID representation improves personalized generation when compared to other face representation using the same personalization approach.

\section{Method}\label{sec:method}

Let $\mathcal{X} = \{ x_1, x_2, \dots, x_M \}$ represent a set of $M$ input images of an individual  where each image $x_i \in \mathbb{R}^{H \times W \times 3}$ depicts the individual’s face under varying poses, expressions, and lighting conditions. 
Our goal is to create a holistic facial identity representation, 
\begin{equation}
    \ell = E(\mathcal{X}), 
\end{equation}
that captures an individual's appearance and its nuanced variations across different contexts, poses, and expressions. 

To this end, we introduce a new face representation  named \methodname, featuring an \methodnametwo~Encoder and a novel \emph{few-to-many identity reconstruction} training with a \emph{multi-decoder} objective.
Designed for generative tasks, this representation aims to enable high-fidelity face generation in diverse poses and expressions, supporting a wide array of generative applications.

In the following, \cref{sec:method:encoder} describes our the \methodnametwo~encoder architecture.
\cref{sec:method:f2m} details the few-to-many identity reconstruction task used during training. 
\cref{sec:method:decoders} discusses the multi-decoder objective, which has two complementary decoding objectives, each applied within the few-to-many identity reconstruction framework.
Lastly, \cref{sec:dataset} introduces the dataset we curated to maximize the potential of the proposed training strategy.

\begin{figure}[t]
    \centering
    \includegraphics[width=0.9\columnwidth]{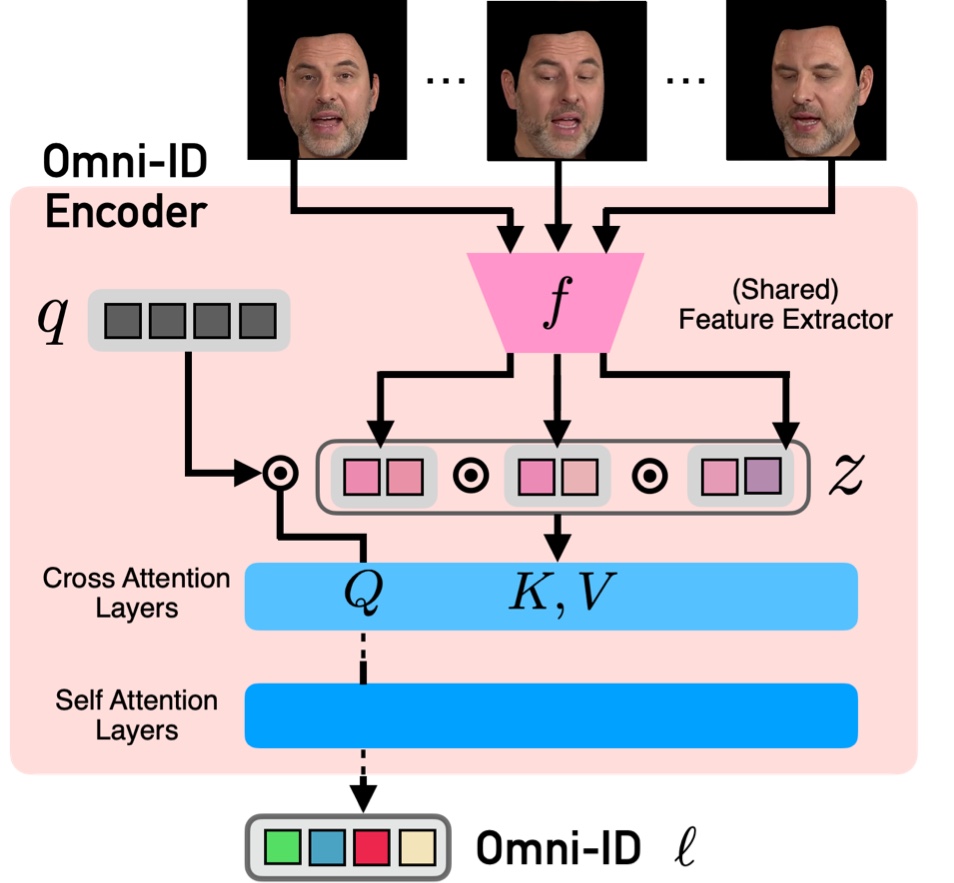}
    \caption{
    \textbf{Omni-ID Encoder} receives a set of images of an individual, projects them into keys and values, which are then fed into cross-attention layers. These layers attend to \protect\hypersetup{linkcolor=cvprblue}\protect\hyperref[fig:attention]{learnable queries}
    that are semantic-aware, allowing the encoder to capture shared identity features across images. 
    Self-attention layers refine these interactions further, producing a holistic representation $\ell$.}
    \label{fig:encoder}
\end{figure}

\subsection{\methodname~Encoder}
\label{sec:method:encoder}

Our proposed \methodnametwo~Encoder $E$ encodes an image set $\mathcal{X}$, with any number of images, into a holistic representation for the identity $\ell = E(\mathcal{X}) \in \mathbb{R}^{L \times C}$.  $L$ represents the token length and $C$ denotes the number of dimensions. 
In order to support encoding an image set, the key design decisions revolve around how to combine individual image features, $ \ell = f(x_i) \in\mathbb{R}^{L_x \times C} $, where $L_x$ is the token length for the image features. 
For this, we use a transformer architecture with a learnable token $q\in \mathbb{R}^{L \times C}$.  
The individual image features are first concatenated in the token-axis to form the image set feature, $z = [\ell_0; \ell_1; ... ;\ell_M] \in\mathbb{R}^{(M\cdot L_x) \times C} $, and then integrated in the encoder through cross-attention layers as keys and values. 
Our full transformer architecture consists of multiple cross-attention layers (all with KV-injection from $z$) followed by multiple self-attention layers. 
See~\cref{fig:encoder} for an overview of our architecture.

\subsection{Few-to-Many Identity Reconstruction}
\label{sec:method:f2m}
During training, given the full set of images of an individual, $\mathcal{X} = { x_1, x_2, \dots, x_N }$, we randomly select a small input subset $\mathcal{X}^s$, and another larger  set $\mathcal{X}^r$ as reconstruction targets, where $|\mathcal{X}^r| > |\mathcal{X}^s|$. 
The model is tasked with utilizing the input subset $\mathcal{X}^s$ for generating the target subset $\mathcal{X}^r$. 

Given the encoder feature $\ell$, each of the decoders $D$ is tasked to conditionally reconstruct all the target images:
\begin{gather}
    \hat{x}^r_i = D(\ell, \tilde{x}^r_i), \quad \forall x^r_i \in \mathcal{X}^r, \\
    \tilde{x}^r_i = \text{Corruption\_Process}(x^r_i),
\end{gather}
where $\tilde{x}^r_i$ is the corrupted target image.  
Intuitively, a corruption process destroys information from the target image, and prevents the decoder from utilizing the target image to achieve autoencoding.  
This forces the decoder to rely on the encoder feature $\ell$ to infer identity information, thereby encouraging the encoder to learn a robust identity representation. 
Meanwhile, the corrupted target image still retains cues about other conditions, such as lighting, pose, and subtle hints of expression, providing essential context that, when combined with representation from the encoder, enables accurate reconstruction. 
To leverage the strengths of different corruption types, we employ distinct objectives, all with few-to-many identity reconstruction, which we detail below.

\subsection{Multi-Decoder Objectives}
\label{sec:method:decoders}

Our multi-decoder objective optimizes a single encoder by $K$ distinct decoders. In our design, we utilize two (i.e., $K$=2) complementary decoding objectives: a conditional masked reconstruction objective referred to as the Masked Transformer Decoder (MTD) and a conditional Flow-Matching objective.
The MTD objective is suitable for learning a representation with wide coverage, but on its own suffers from neglecting fine-grained details.  
The Flow-Matching objective excels to picking up fine-grained details by training at various noise levels, but on its own is less effective for representation learning (shown in~\cref{sec:ablations}). 

\begin{figure}[t]
    \centering
    \includegraphics[width=\columnwidth]{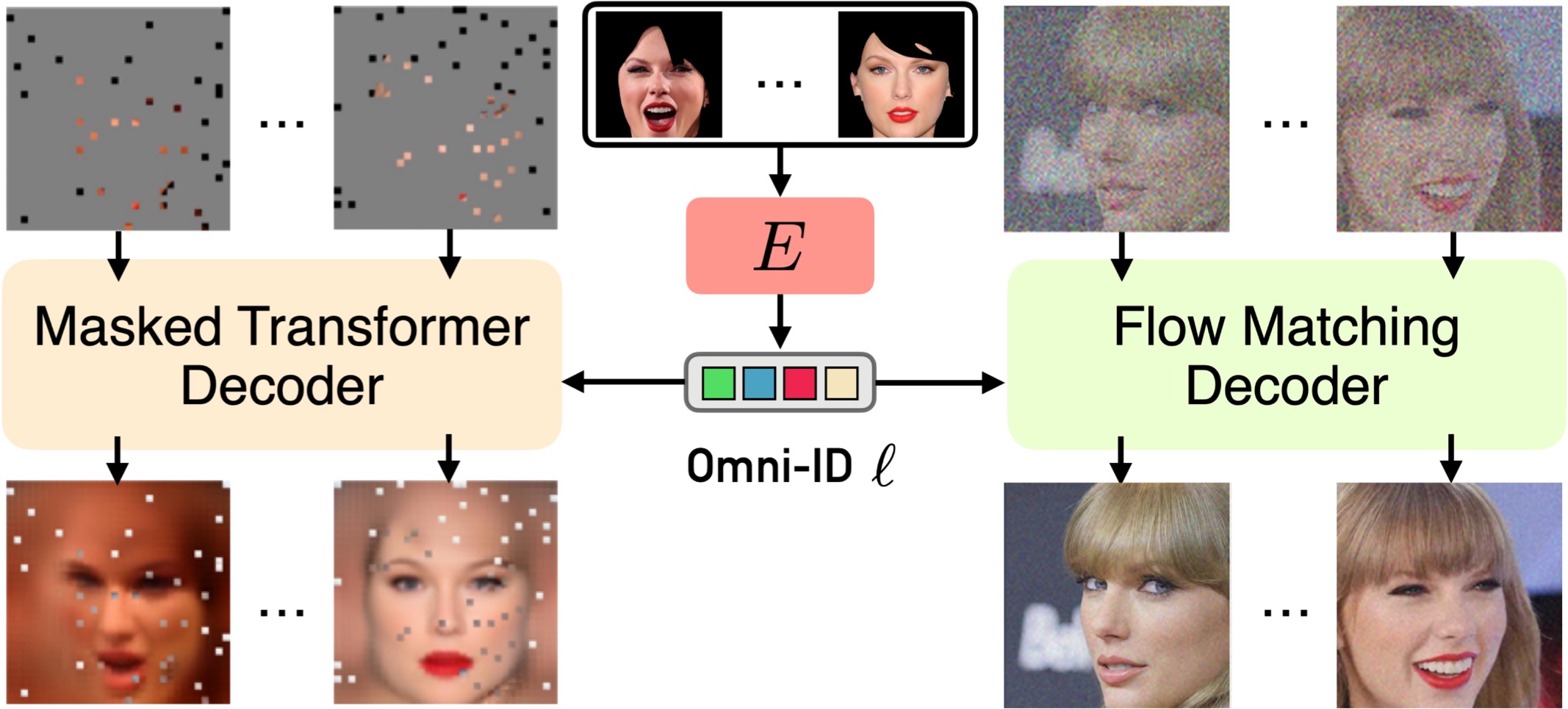}
    \caption{\textbf{Multi-decoder training}. 
    (left) Masked Transformer Decoder (MTD) is designed to reconstruct unseen facial pixels from the Omni-ID representation and a minimal subset of visible pixels which do not leak identity.
    (right) Flow Matching Decoder enhances the encoder by a higher-quality reconstruction task. 
    }
    \label{fig:decoders}
\end{figure}

\paragraph{Decoder 1: Masked Transformer Decoder}
Our first decoding objective is a variant of conditional masked autoencoding we call \textit{Masked Transformer Decoder} (MTD), where the decoder receives as inputs the \methodname~representation $\ell$, and a heavily masked version of targets to reconstruct (i.e. $95\%$ of tokens masked).  
By applying a very high masking ratio, MTD ensures subject’s anonymity and that the identity information is solely derived from $\ell$.  
The minimal visible/unmasked pixels provide essential contexts such as pose and lighting.
The \methodnametwo~Encoder and the Masked Decoder $D$ are trained end-to-end using a reconstruction loss as follows:
\begin{align}\label{eqn:mtd}
\mathcal{L}_1 &= \frac{1}{|\mathcal{X}^r|} \sum_{x^r \sim \mathcal{X}^r}\Big|  \big(D (\ell , \tilde{x}^r 
 )  - x^r \big) \odot M^r \Big|_1, \\
 &\ell = E(\mathcal{X}^s), \;\;\;\; \tilde{x}^r=x^r \odot M, 
\end{align}
where $M$ is the randomly sampled mask to corrupt the target image $x^r$ and $M^r$ is the face segmentation mask to remove background.
Refer to \figlabel\ref{fig:decoders} (\emph{left}) for a practical example of how inputs and masked prediction appear during MTD training.
The decoder uses the same architecture as the \methodname~encoder, but instead of a learned query, its query comes from the masked input target image $\tilde{x}^r$. The identity feature $\ell$ is  fed through the cross attention layers and serves as keys and values in the decoder.

\begin{table*}[t]
\centering
\caption{\textbf{Quantitative comparisons to different representations} on controllable face generation. The backbone is the same for all methods: IP-Adapter + ControlNet. All baselines undergoes Flow-Matching pretraining to initialize IP-Adapters to converge for fair comparison. We show results of using different number of inputs in two test sets.
}
\label{tab:posed_face}
\begin{tabular}{@{}cccccc@{}}
\toprule
Dataset & \multicolumn{2}{c}{\textbf{MFHQ Test}} &\multicolumn{2}{c}{\textbf{Webface Test}~\cite{Arc2Face}} \\ 
Metrics & \textbf{ID Similarity}$\uparrow$ & Pose Error$\downarrow$ & \textbf{ID Similarity}$\uparrow$ & Pose Error$\downarrow$\\ 
\#inputs &  1 / 3 / 5 / 7 & 1 / 3 / 5 / 7 & 3 / 5 / 8 / 16 & 3/ 5 / 8 / 16  \\ 
\midrule
ArcFace & 0.515 / 0.523 / 0.529 / 0.535 & 2.4/2.3/2.3/2.3 & 0.379 / 0.373 / 0.370 / 0.371 / 0.373 & 2.5/2.4/2.5/2.4/2.4\\
CLIP & 0.648 / 0.670 / 0.680 / 0.682 & 2.3/2.3/2.2/2.2 & 0.695 / 0.696 / 0.696 / 0.695 / 0.695 & 2.9/2.8/2.8/2.7/2.8\\ 
ArcFace + CLIP & 0.638 / 0.655 / 0.663 / 0.664 & 2.4/2.4/2.3/2.3 & 0.652 / 0.654 / 0.656 / 0.658 / 0.655 & 2.9/2.8/2.8/2.7/2.8\\
\textbf{Omni-ID (Ours)} & \textbf{0.708} / \textbf{0.728} / \textbf{0.737} / \textbf{0.742} & 2.1/2.0/2.0/2.0  & \textbf{0.774 / 0.779 / 0.781 / 0.784 / 0.785} & 2.7/2.6/2.6/2.6/2.6\\

\bottomrule
\end{tabular}%
\end{table*}

\paragraph{Decoder 2: Conditional Flow Matching}

The MTD objective, being a variant of autoencoding, serves as a effective approach for learning a wide covering representation.
However, it suffers from the pitfalls of an autoencoding objective, which tends to produce blurry outputs and omit fine-grained details. 
To capture more nuanced details, a decoder that is able to recover nuanced details is required. 
For this, we resort to diffusion decoders in conditional flow matching.
These decoders are optimized to remove noise from a noisy target at various noise-levels encouraging our model to learn details at all noise-levels. 

The \methodnametwo~Encoder and the diffusion decoder $V$ are optimized jointly by flow matching objective: 
\begin{align}
\mathcal{L}_2 &= \mathbb{E}_{x^r \sim \mathcal{X}^r, t, \epsilon} \left[ \left\| V\left(\tilde{x}^r_{t,\epsilon} , t, y, \ell \right) - (\epsilon - x^r) \right\|_2^2 \right], \\
 &\ell = E(\mathcal{X}^s), \;\;\;\; \tilde{x}^r_{t,\epsilon} = (1 - t)x^r + t\epsilon, 
\end{align}
where  $t \sim \mathcal{U}(0, 1) $ is the time step and $\epsilon \sim \mathcal{N}(0, I) $ is a noise sample from the standard normal distribution and $y$ is a fixed text prompt (\texttt{"photo of a person."}). 
Refer to \figlabel\ref{fig:decoders} (\emph{right}) for an example of how inputs and targets appear during Flow-Matching training.
Decoder $V$ is a combination of a pretrained flow model~\cite{FLUX} and IP-Adapter \cite{ye2023ip-adapter}.
Similarly to IP-Adapter, we project \methodname~representation into keys and values and inject them via learnable decoupled attention layers into the pretrained flow decoder.

\subsection{MFHQ Dataset} 
\label{sec:dataset}

\methodname~training requires a large-scale dataset with many identities, each with multiple face images. The closest existing datasets that meet this requirement are the ones used for face recognition, \eg WebFace42M \cite{zhu2021webface260m}. However, the quality of these datasets is insufficient to train generative face representations due to two major limitations. 
First, they are low-resolution (typically $112$$\times$$112$), and the representations trained on the up-sampled versions of them tend to smooth out the fine-grained details \cite{Arc2Face}. 
Second, intra-identity variations in face recognition datasets are usually too high due to age and quality variations, making the generative representation trained on them unable to encode a consistent facial identity. Refer to \cref{sec:ablations} for examples. 

We thus introduce a new large-scale dataset \textbf{MFHQ}--\textbf{m}ultiple \textbf{f}aces in \textbf{h}igh \textbf{q}uality. 
MFHQ consists of 134,077 identities with 8 images per ID collected from videos to ensure identity consistency. The face resolution is filtered to be larger than $448$. MFHQ overclusters the video frames based on their estimated head poses, then samples 8 faces for each ID according to the face quality estimation \cite{chen2024topiq}. This clustering-based sampling ensures pose differences. 
The video sources come from a combination of CelebV-HQ~\cite{CelebVHQ}, VFHQ~\cite{VFHQ}, TalkingHead-1KH~\cite{wang2021facevid2vid}, and CelebV-Text~\cite{CelebV-Text}. MFHQ collection is illustrated in \supp.

\section{Experiments}\label{sec:exp}

In this section, we validate the learned \methodname~representation by evaluating its performance on two downstream generative tasks. In both tasks, we compare the \methodname~representation to existing identity representations, demonstrating improved identity fidelity and adaptability.

The first task is \emph{controllable face generation} (\cref{sec:exp:posedface}), where a downstream generator produces an image of an individual in unseen poses based on an identity representation and a target pose (i.e., landmarks). This task tests the representation’s ability to capture nuanced changes with varying poses and expressions.
The second task is \emph{personalized text-to-image generation} (\cref{sec:exp:personalization}). Here, the generator creates scene-level images that maintain both individual identity and the quality of the original text-to-image model. 
Lastly, we validate our design choices including each of the few-to-many identity reconstruction training paradigm, the decoding objectives, our proposed dataset, and their hyperparameters (\cref{sec:ablations}).

\subsection{Implementation Details}
\noindent\textbf{\methodname~encoder.}
Our Omni-ID encoder uses CLIP-H \cite{CLIP} as the feature extractor and finetunes all layers. Omni-ID encoder uses a learnable query with $L$=$256$, $C$=$1280$ and $2$ cross-attention blocks and $2$ self-attention blocks, which is sufficient to learn a representation from image features.

\begin{figure*}[t] 
    \centering \includegraphics[width=1.0\textwidth]{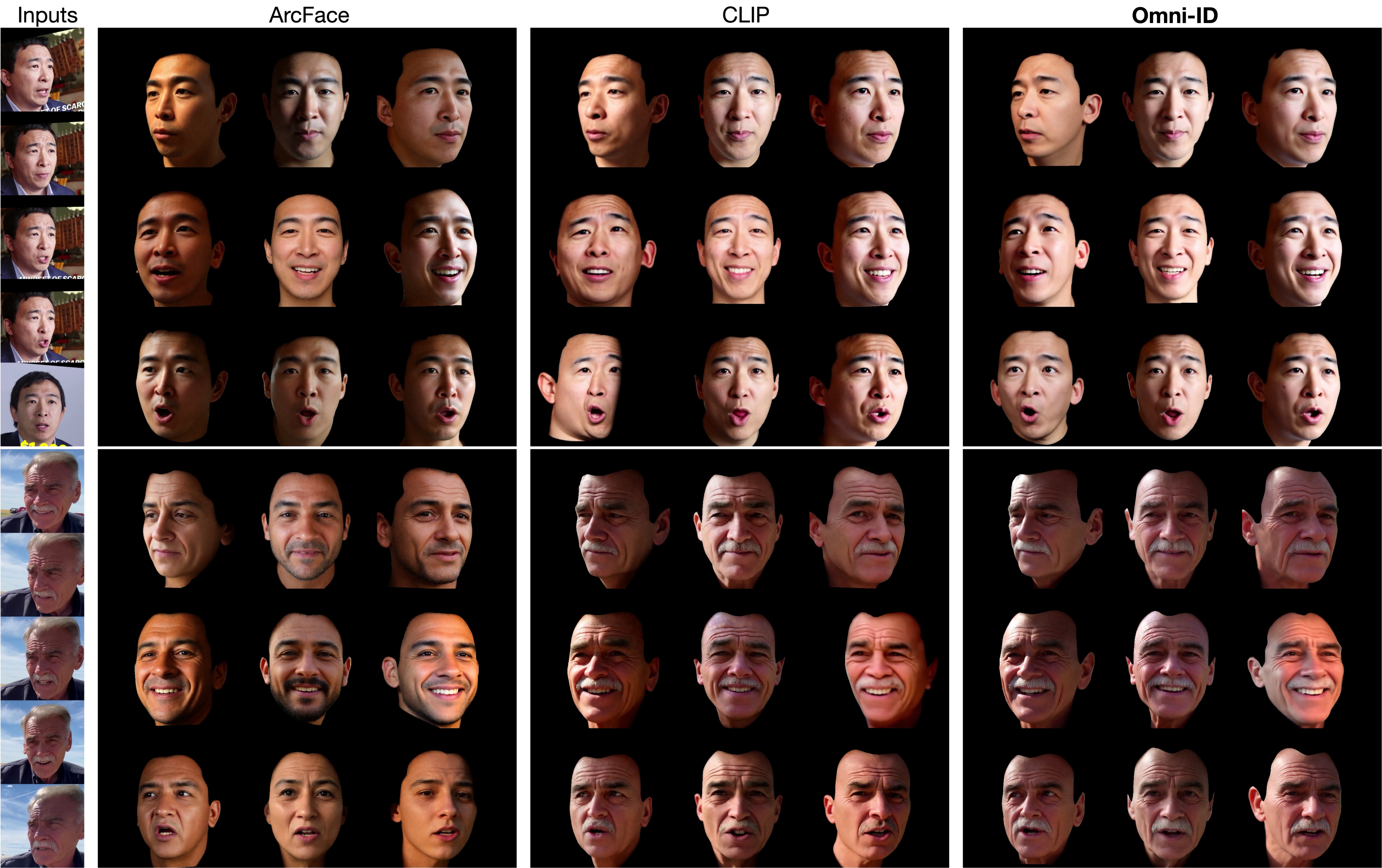}
    \caption{\textbf{Qualitative comparisons in controllable face generation.} We train the same IP-Adapter+ControlNet for each representation. 
    \methodname~achieves superior identity preservation and captures nuanced changes with varying poses and expressions more faithfully.}
    \label{fig:sota-face}
\end{figure*}

\noindent\textbf{\methodname~training.}
Our MTD decoder is trained for $200$K steps with a masking ratio of $95\%$.  
Our Flow-Matching Decoder is trained for $10$K steps using FLUX dev \cite{FLUX} as the base model. 
Both stages are trained in MFHQ with $44$ held out videos as testing and others as training. 

\noindent\textbf{Baseline representations. } In the controllable face generation task, we compare our \methodname~representation to other commonly used identity representations: pretrained CLIP features~\cite{CLIP} and ArcFace embedding~\cite{ArcFace}.  
We also compare to CLIP+ArcFace, following FaceIDPlus \cite{ye2023ip-adapter}, where the ArcFace embeddings are projected into queries and CLIP features are used as keys and values to get the representation through the same attention mechanism as our Omni-ID. For CLIP representations, we use all $257$ tokens for all baselines. For ArcFace, we project embedding $\mathbb{R}^{1\times 512}$ to $256$ tokens $\mathbb{R}^{256\times 512}$ for better quality.
For both tasks, we ensure fair comparisons with the baselines  (see \supp).

\noindent\textbf{Metrics.}
We use ID similarity as metric, which is measured using the cosine similarity between the face recognition features \cite{facenet-pytorch}  extracted from the generation and ground truth. In controllable face generation, 
pose error \cite{HopeNet} metric is also provided which is measured by the sum of absolute differences of yaw, pitch, roll in degrees.

\begin{figure}[t] 
\centering
\includegraphics[width=1.0\linewidth]{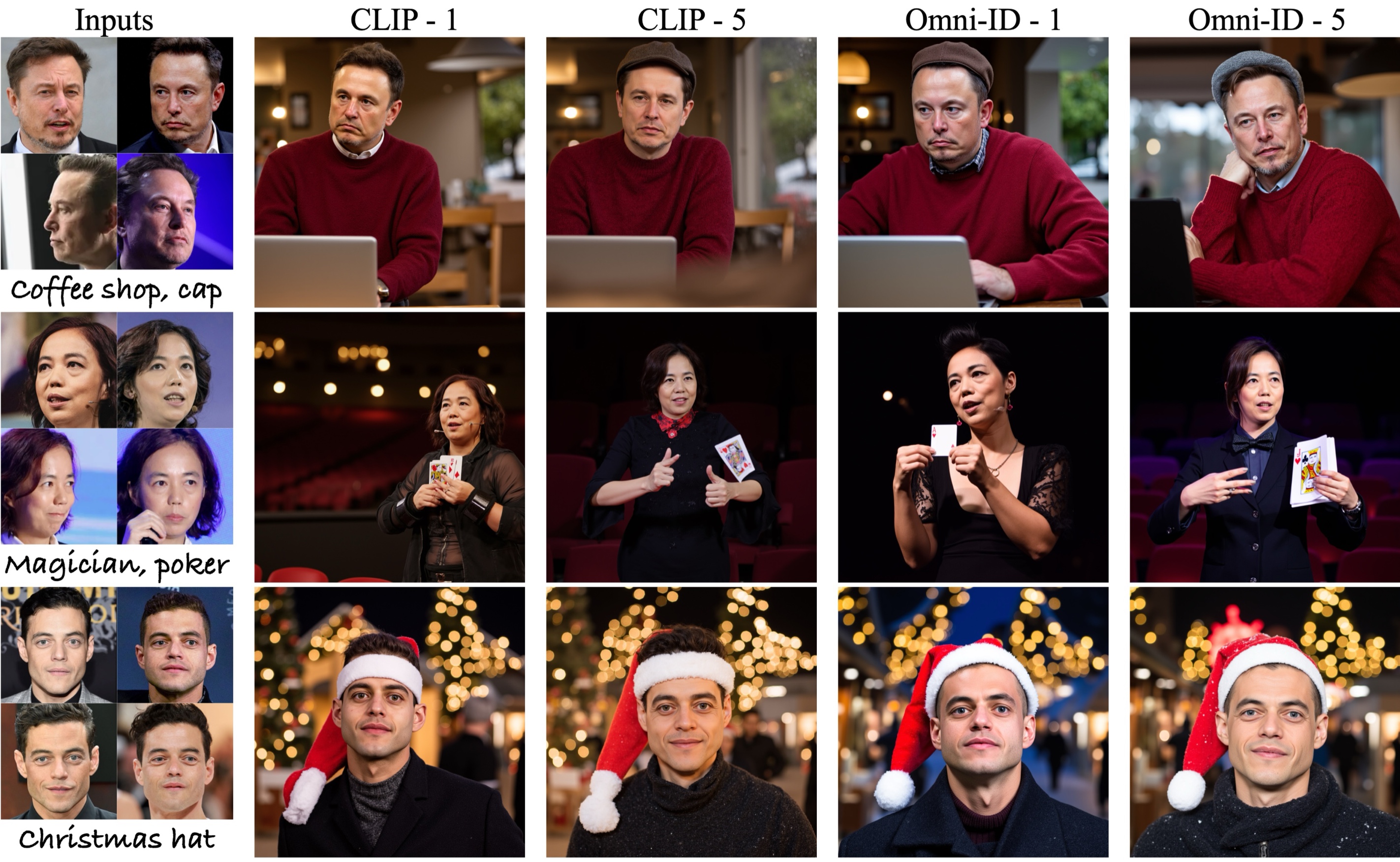}
\caption{\textbf{Qualitative comparisons with different representations in personalized T2I generation.} We show results of the same IP-Adapter trained with different representations. \methodname~achieves better ID preservation for both single and multiple input images. See more examples and how Omni-ID significantly outperforms other personalization methods \cite{wang2024instantid, li2023photomaker} in \supp. 
}
\label{fig:sota-person}
\end{figure}

\subsection{Controllable Face Generation}
\label{sec:exp:posedface}

Given a pretrained representation, we train a combination of ControlNet~\cite{ControlNet} and IP-Adapter with frozen FLUX for controllable face generation. ControlNet receives as input target landmark pose, where IP-Adapter injects frozen face representations. 
The methods are evaluated in two test sets: (1) all identities from Webface21M \cite{zhu2021webface260m, Arc2Face} consisting of at least 16 photos with minimum ID similarity 0.6 and minimum pose differences $7\degree$, and (2) MFHQ test set. 
Compared to ArcFace, CLIP, or their combined representations, \methodname~shows better performance in identity preservation in ~\cref{tab:posed_face}. All methods achieve similar pose errors indicating their convergence. Beyond metrics,~\cref{fig:sota-face} highlights qualitative differences between representations using 5 inputs driven by template landmarks. Although ArcFace encodes facial features effectively for recognition tasks, it is overly invariant to attributes such as age and skin tone. CLIP preserves general visual features but struggles with adaptivity to new poses and expressions due to its instance-level encoding and lack of fine-tuning for facial details. Consequently, facial features such as beards (last row) are not accurately represented in CLIP, and sensitivity to pose and expression changes is noticeable. In contrast, \methodname~achieves high-fidelity identity preservation, capturing facial details across diverse poses and expressions.

\begin{figure}[t]
\centering
\includegraphics[width=1.0\linewidth]{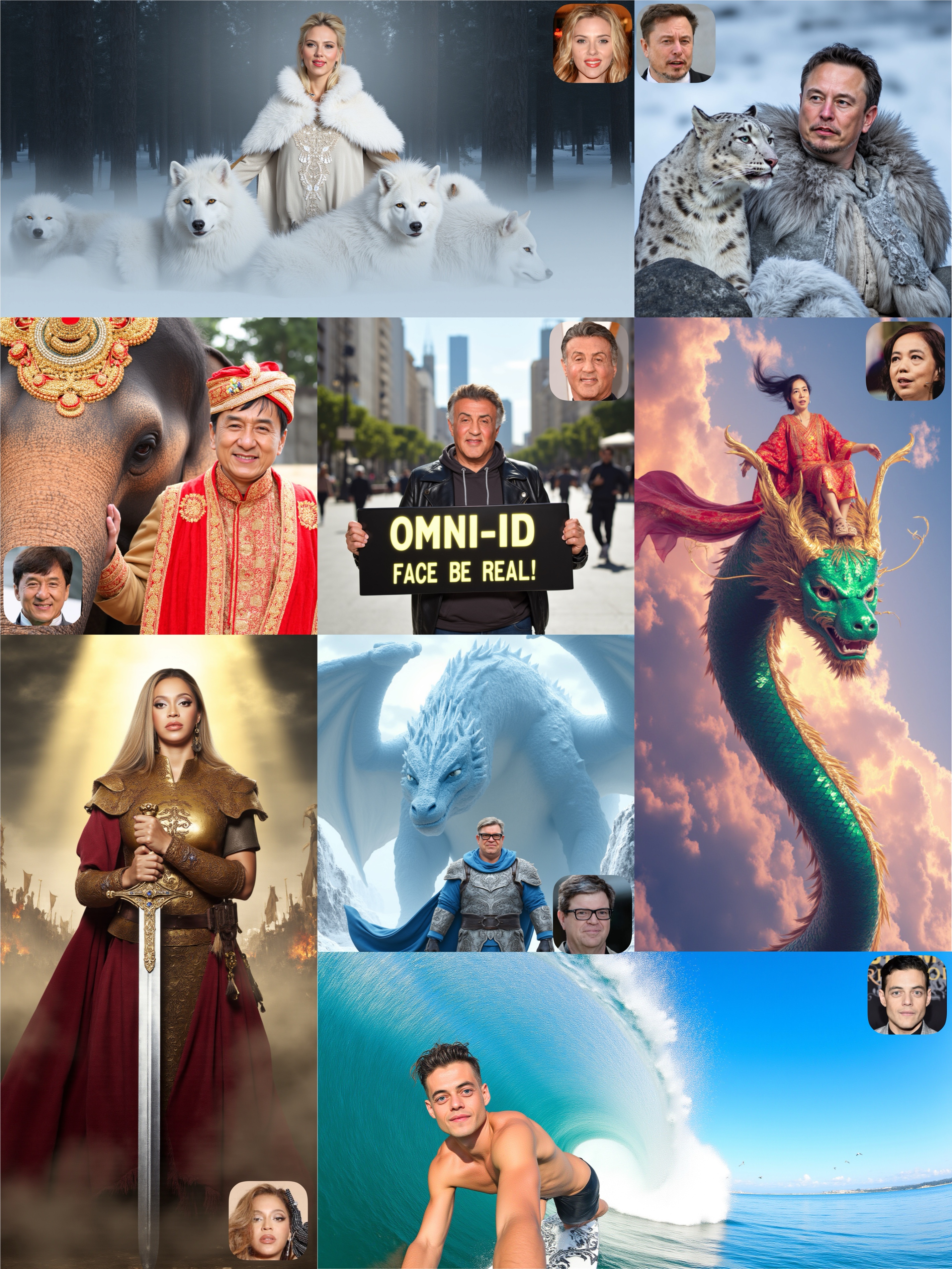}
\caption{\textbf{Gallery of Omni-ID in personalized T2I generation.} Omni-ID enables high identity preservation. Results achieved by injecting Omni-ID representation through IP-Adapter~\cite{ye2023ip-adapter} into the frozen FLUX dev model~\cite{FLUX} without LoRA~\cite{hu2022lora} or postprocessing.
}
\label{fig:gallery}
\end{figure}

\subsection{Personalized T2I Generation}
\label{sec:exp:personalization}

Given a pretrained representation, we train an IP-Adapter to inject into frozen FLUX-dev. Notice, this differs from the denoising decoder we used during representation learning in its data. Here, the input data is an image of the face, but the target image is a scene-level image and has a corresponding text caption. We train all baselines in the same internal licensed image dataset ($\sim$1M single-view images), and evaluate them on $10$ identities and $20$ diverse prompts. ID similarity is employed as the metric. 
See \supp for the quantitative results. As can be seen in~\cref{fig:sota-person}, Omni-ID representation demonstrates superior performance in terms of identity preservation when applied to personalized text generation, outperforming CLIP in both single input image, as well as in the the multiple input images case.
See \cref{fig:gallery} for more results of Omni-ID. See \supp for more qualitative comparisons using different base models \eg SD \cite{ldm}.

\subsection{Ablations \& Analyses}
\label{sec:ablations}

\begin{table*}[t]
\centering
\caption{\textbf{Validating MTD decoder and its design decisions.} This table summarizes the face generation quality from the Flow-Matching Decoder described in~\cref{sec:method:decoders} with varied configurations in the MTD pre-training. 
For each configuration, we report results using 1 or 3 input image(s) (1-image / 3-image). `I-O' denotes the number of input and output images used in training.
}
\label{tab:ablate_mtd}
\resizebox{0.88\textwidth}{!}{%
\begin{tabular}{@{}l|c|c|ccc|cc@{}}
\toprule
\textbf{Ablation} & \textbf{Ours full} & \textbf{w/o MTD} & \multicolumn{3}{c|}{\textbf{Few-to-many}} & \multicolumn{2}{c}{\textbf{MTD mask ratio}} \\ 
I-O, mask ratio & 3-8, 0.95 & --- & 3-1, 0.95 & 3-5, 0.95 & 8-8, 0.95 & 3-8, 0.99 & 3-8, 0.85 \\ 
\midrule
\textbf{ID Similarity $\uparrow$} & \textbf{0.683 / 0.733} & 0.336 / 0.358 & 0.491 / 0.515 & 0.582 / 0.615 & 0.661 / 0.696 & 0.670 / 0.700 & 0.609 / 0.650 \\ 
\bottomrule
\end{tabular}%
}
\includegraphics[width=0.88\textwidth]{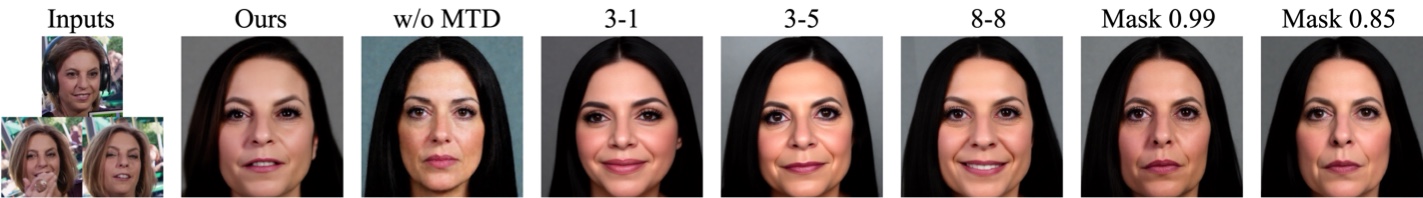}
\vspace{-1em}
\end{table*}

\cref{tab:ablate_mtd} ablates MTD pre-training configurations through face generation qualities by the Flow-Matching decoder.
\cref{tab:ablate_diffusion} ablates pre-training objective and dataset in the task of downstream controllable face generation. We present results using 1 or 3 input images in the ablation study, as additional inputs yield only marginal performance improvements as aforementioned and the models are trained with at most 3 inputs.

\noindent\textbf{Few-to-many identity reconstruction.}
\cref{tab:ablate_mtd,tab:ablate_diffusion} demonstrates the few-to-many reconstruction task is better than the conventional alternative of single-image reconstruction.
In \cref{tab:ablate_mtd}, we observed increasing performance as we increase the number of target images (compare 3-8 with 3-1 and 3-5). Note 3-8 outperforms 8-8 since the latter might reconstruct all inputs, whereas 3-8 is always optimized to also reconstruct unseen images with new poses and expressions. 
In \cref{tab:ablate_diffusion}, the performance significantly degrades when using single-image reconstruction for both decoding objectives (`$-$ Few-to-many pretraining' row). 

\begin{figure}[t] 
\centering
\includegraphics[width=0.47\textwidth]{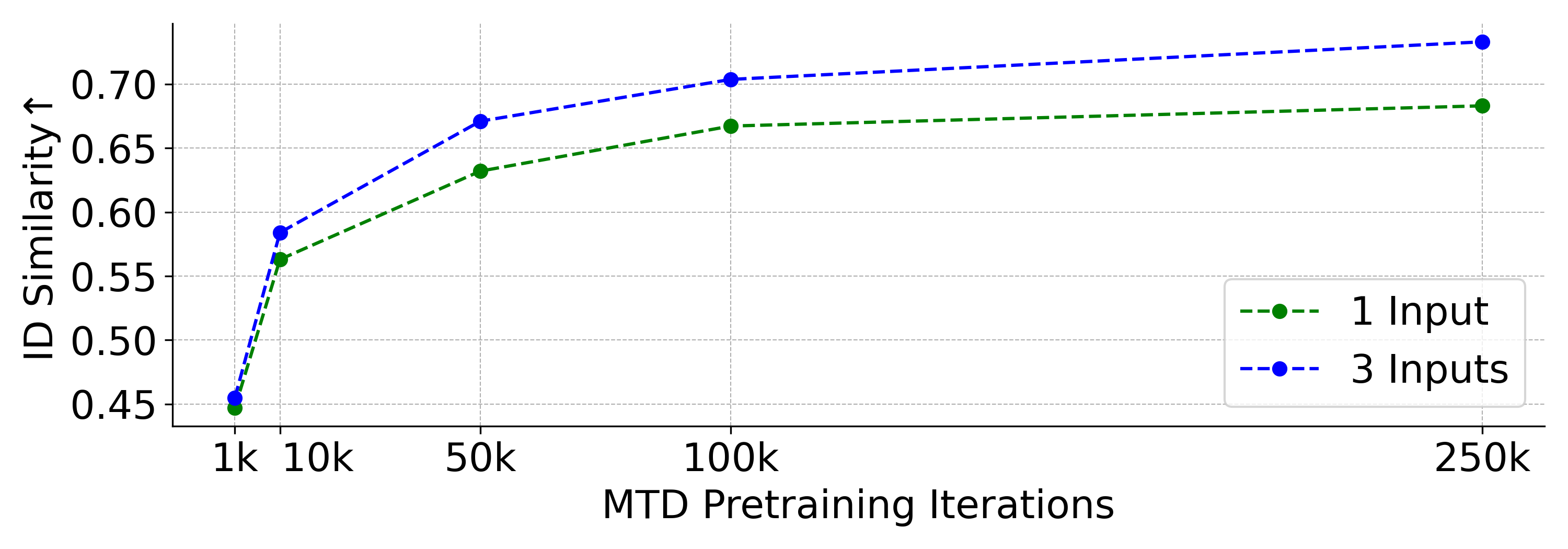}
\includegraphics[width=0.47\textwidth]{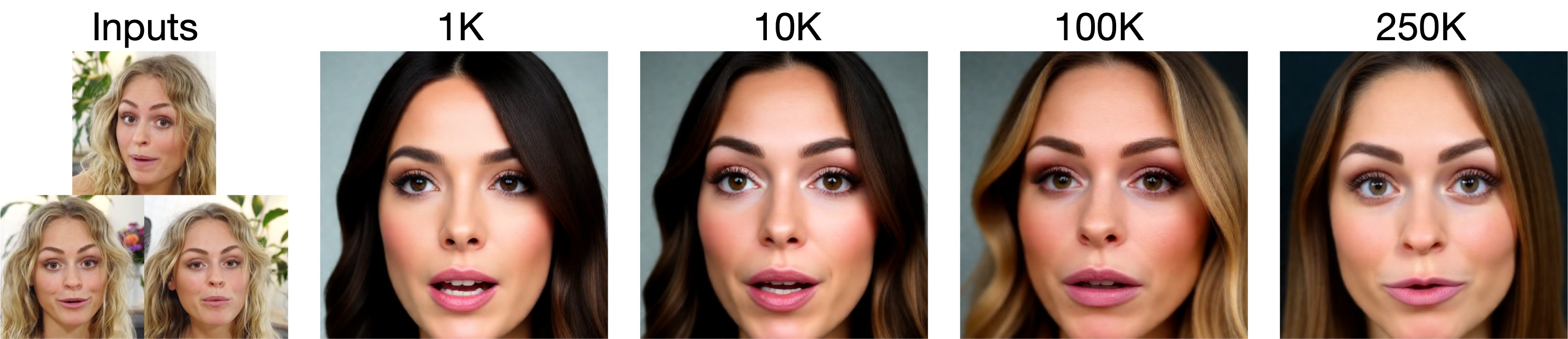}
\caption{
\textbf{More MTD pretraining consistently improves ID preservation}. That curve shows the quantitative results from 1 or 3 inputs with increasing \textit{MTD pretraining} steps followed by the same Flow-Matching Decoder training steps for fair comparisons. 
}
\label{fig:mtd_curve}
\end{figure}

\noindent\textbf{MTD objective.} 
In \cref{tab:ablate_mtd}, MTD pre-training results in better performance with generally a higher masking ratio.  
However, as the masking ratio reaches $99\%$, the performance drops slightly.
Intuitively, MTD benefits from a high masking ratio, but too high of a masking ratio also makes the reconstruction task ill-posed and noisy.  
In addition, \cref{fig:mtd_curve} demonstrates MTD pretraining is beneficial for the Flow-Matching decoder training consistently at almost any number of steps and consistently improves encoding.
Lastly, \cref{tab:ablate_diffusion} shows that removing MTD pre-training harms the downstream controllable face generation.

\noindent\textbf{Flow-Matching objective.}
As shown in~\cref{tab:ablate_diffusion}, removing the Flow-Matching decoding objective leads to a lower identity similarity score and a noticeable loss of fine-grained details (e.g., less defined beards and smoother faces as shown in the figure).
In the Flow-Matching decoder, due to the different noise-levels, it encourages the representation to encode the fine-grained details. 

\begin{table}[t]
\centering
\caption{\textbf{Ablate Flow-Matching Decoder training} evaluated in controllable face generation. Both Flow-Matching Decoder pretraining and MFHQ dataset enhances details. Few-to-many identity reconstruction training and MTD improve ID preservation. 
}
\label{tab:ablate_diffusion}
\resizebox{\columnwidth}{!}{
\begin{tabular}{@{}l|cc@{}}
\toprule
\textbf{Ablation} & \textbf{ID Similarity} $\uparrow$ & \textbf{Pose Error} $\downarrow$ \\ 
\midrule
Ours full & \textbf{0.708} / \textbf{0.728} & 2.1 / 2.0 \\
\cdashline{1-3}
$\;\;-$ MTD pretraining & 0.468 / 0.473 & 2.8 / 3.1 \\ 
$\;\;-$ Flow-Matching Decoder pretraining & 0.672 / 0.685 & 2.4 / 2.3 \\ 
$\;\;-$ Few-to-many pretraining & 0.616 / 0.633 & 2.3 / 2.2 \\
$\;\;-$ Pretraining on MFHQ & 0.678 / 0.693 & 2.2 / 2.1 \\ 
\bottomrule
\end{tabular}%
}
\centering
\includegraphics[width=\columnwidth]{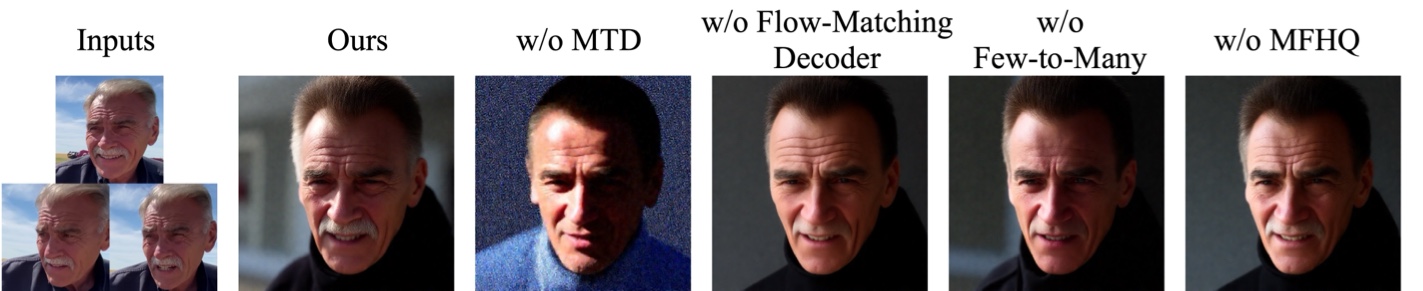}
\end{table}

\noindent\textbf{Effectiveness of MFHQ.}
Results in \cref{tab:ablate_diffusion} validate the utility of our MFHQ dataset.  When we replace our training data with the existing alternative, WebFace21M~\cite{Arc2Face}, we see a drop in the identity fidelity.  This is due to the larger intra-class ID variation, which introduces noise in training.

\begin{figure}[t] 
\centering
\includegraphics[width=0.97\linewidth, trim=0 175 0 0, clip]{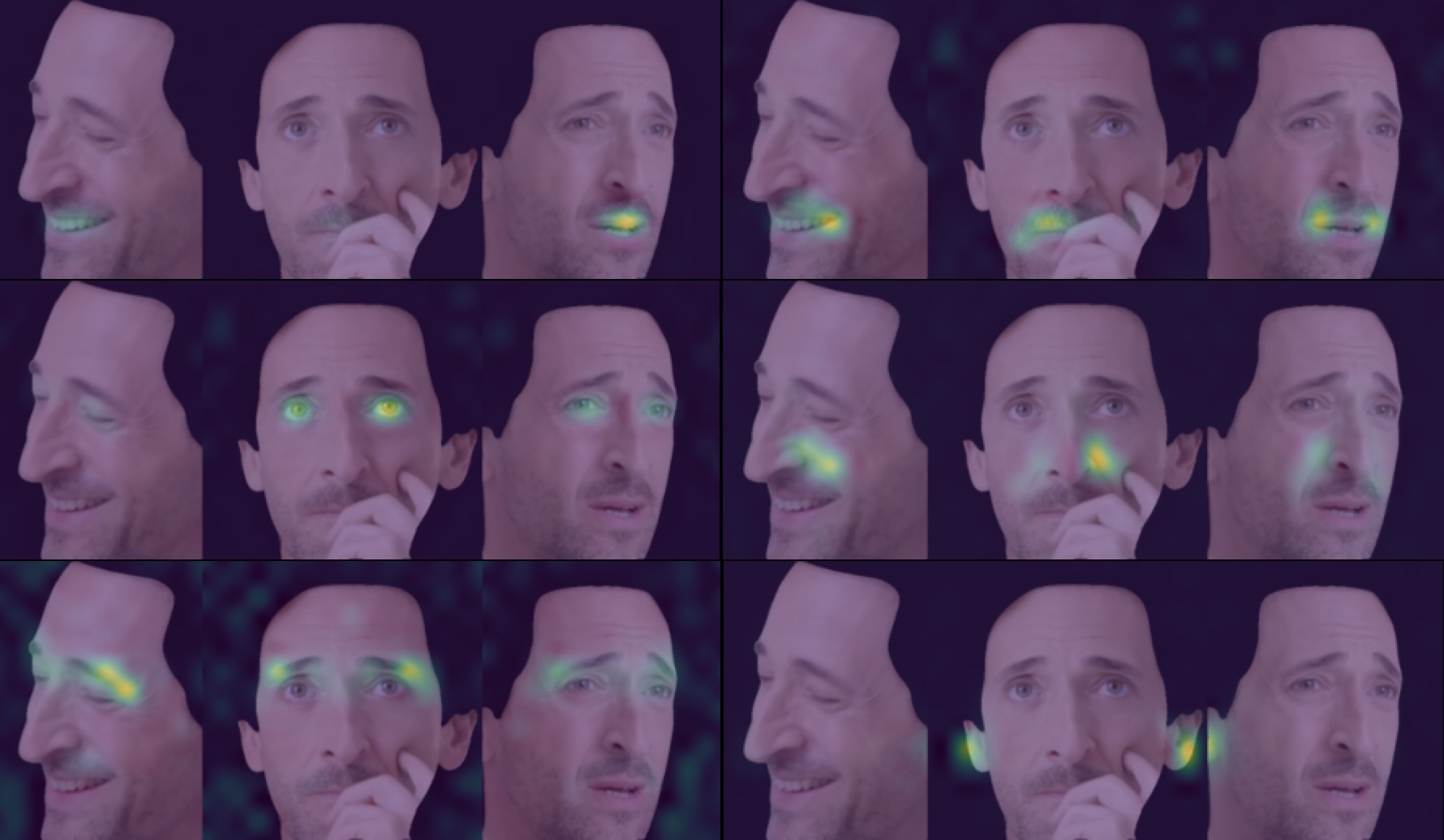}
\caption{\textbf{Visualization of attention maps} between individual learned query and the keys extracted from input images. Notably, different queries focus on distinct semantic and specific regions of the face. The learned queries also effectively adapt to variations in input facial features, such as open or closed mouths and eyes, as well as to occlusions like hands or missing features, such as ears.
}
\label{fig:attention}
\end{figure}

\noindent\textbf{Attention visualization.} We visualize the attention maps of our \methodname~ encoder in \cref{fig:attention}. Notably, the same learned token attends to different patches across various input images based on semantic context. For instance, the same query feature results in a different attention map depending on whether the eyes are open or closed, while queries focused on the mouth region ignore a hand occluding it. These results demonstrate that Omni-ID learns to consolidate visual information scattered across an unstructured set of input images into a structured representation, where each entry represents certain global or local identity features.

\section{Conclusions}\label{sec:con}
We introduced Omni-ID, a facial representation tailored for generative tasks, which captures an individual's holistic appearance across various expressions and poses. Trained in a few-to-many identity reconstruction framework with a multi-decoder objective, Omni-ID encodes a fixed-size tokenized representation from diverse unstructured input images, demonstrating superior identity preservation. Unlike discriminative representations like ArcFace and CLIP, Omni-ID retains nuanced identity information critical for high-fidelity generation. Moreover, the quality of Omni-ID improves with an increasing number of input images.

Our results suggest that generative identity representation holds transformative potential for diverse facial generation applications.
We anticipate that our approach will inspire further innovation, broadening the capabilities and scope of generative identity modeling across a wider range of applications.
Improvements in dataset scale and consistency, as well as the number and type of the decoders would further enhance robustness. Collection of large-scale multiple photos of the same identity in a more diverse context or lighting augmentation will also improve the robustness of Omni-ID, addressing lighting injection and skin tone predominance issues. Additionally, Omni-ID does not represent attributes that are not intrinsic to the face, such as hair, which can result in these features being “hallucinated” in downstream tasks. Extending Omni-ID to include a more comprehensive set of attributes remains an open direction.

\inlinesection{Acknowledgement}\label{sec:ack}
The authors would like to acknowledge Jian Wang, Qiang Gao, and Sizhuo Ma from the Computational Imaging team at Snap for their advices with face quality assessment. We also thank other members of the Snap Creative Vision team for their valuable feedback and insightful discussions throughout the course of this project.

{
    \small
    \bibliographystyle{ieeenat_fullname}
    \bibliography{main}
}
\clearpage \clearpage
\appendix

\renewcommand{\thesection}{\Alph{section}}
\renewcommand{\thetable}{\Roman{table}}
\renewcommand{\thefigure}{\Roman{figure}}
\setcounter{section}{0}
\setcounter{table}{0}
\setcounter{figure}{0}

\section{Experiment Details}

\subsection{Baseline Details}

\noindent\textbf{CLIP.} Throughout our experiments, we use CLIP-H from OpenAI \cite{CLIP} as the feature extractor, which works slightly better than CLIP-B/L.
We use the full representation, \ie $257$ tokens ($256$ spatial tokens with $1$ class token) from the second last layer following IP-Adapter Full~\cite{ye2023ip-adapter}, which improves ID preservation compared to using only class token or a reduced number of tokens (\eg $16$). For multiple inputs, we use token concatenation following IP-Adapter, which outperforms simple averaging. 
For fair comparisons, we train IP-Adapter with CLIP representation in the same Flow Matching Decoder stage for 5K steps with an effective batch size $32$ (roughly $1$ epoch in MFHQ). The convergence happens at around $4$K steps.

\noindent\textbf{ArcFace.} We use the ArcFace \cite{ArcFace} model from insightface \cite{insightface} throughout the experiments. 
We project Arcface embedding from $\mathbb{R}^{1\times 512}$ to 256 tokens with $1280$ channels $\mathbb{R}^{256\times 1280}$, which is comparable to CLIP and Omni-ID in terms of representation size. Using $256$ tokens improves its ID preservation compared to using $4$ or $16$ tokens only in IP-Adapter FaceID \cite{ye2023ip-adapter}, and outperforms other reduced number of tokens ($64$).  We concatenate representations in the token dimension for multiple inputs, where averaging merging also reaches a similar results for ArcFace representation. 
We train IP-Adapter with ArcFace representation in the Flow Matching Decoder stage by $75$K steps to converge. Compared to CLIP and Omni-ID which take about $5$K steps, the convergence of ArcFace is rather slow, due to its over-compactness for generative tasks.

\noindent\textbf{ArcFace+CLIP.} Following IP-Adapter FaceIDPlus \cite{ye2023ip-adapter}, Arc-Face+CLIP baseline projects ArcFace tokens from the average ArcFace embeddings in $\mathbb{R}^{1\times 512}$ to $256$ queries in $\mathbb{R}^{256\times 1280}$, where each individual CLIP features in $\mathbb{R}^{257\times 1280}$ are used as keys and values to aggregate multiple inputs. For a fair comparison to Omni-ID, the same transformer with self attention layers is used to merge features. Both the improved number of tokens and the self-attention layers in transformer improves the face quality compared to the original implementation in FaceIDPlus, where $4$ or $16$ tokens are used as query and Q-former \cite{BLIP} without self attentions are employed.

\subsection{Decoders and Training Details}
\paragraph{Decoders details.}
\begin{itemize}
    \item \textit{Masked transformer decoder.} MTD is built by $6$ CA blocks and $2$ SA blocks, which reaches high-quality reconstruction while smaller number of decoder blocks might compensate encoder quality due to the lower decoding ability. Mask ratio of MTD is set to $95\%$, \ie 5\% patches are visible during training, which leads better encoder performance in downstream tasks than mask ratio $85\%$ or $99\%$.  
    The patch size for the decoder is set to $14\!\times\!14$ to balance the speed and quality.   \\
    
    \item \textit{Flow-matching denoising decoder.} For the Flow-Matching Decoder, FLUX dev \cite{FLUX} serves as the base model. We implement a FLUX-based version of IP-Adapter \cite{ye2023ip-adapter}, where the Omni-ID representation is injected into all blocks, including MM-DiT and DiT blocks, via learnable decoupled attention layers. Injecting into both block types results in slightly better quality compared to injecting into only MM-DiT blocks or only DiT blocks, although this improvement is not critical. Each decoupled attention layer optimizes a single linear projection to map $\ell$ from $\mathbb{R}^{L \times C}$ to $\mathbb{R}^{L \times 3250}$, where $3250$ is the channel size used in FLUX. During the Flow-Matching Decoder stage, the Omni-ID encoder and the projection layers of the decoupled attention layers are optimized, while the original parameters in FLUX remain frozen.  \\
    
\end{itemize}

\noindent\textbf{Training Details.}
\methodname~uses a two-stage few-to-many identity reconstruction training process: the MTD stage and the Flow Matching Decoder stage. The MTD stage is trained on our MFHQ dataset at an image resolution of $448$ using a constant learning rate of $1e^{-4}$, an effective batch size of $256$ (distributed as $32$ batches across $8$ NVIDIA A100 GPUs), and the AdamW optimizer for 250K iterations. The Flow Matching Decoder stage is trained on the same dataset at a resolution of $512$, with a constant learning rate of $1e^{-4}$, an effective batch size of $32$, and the AdamW optimizer for 5K iterations. In both stages, we uniformly sample a variable number of inputs ($1$ to $3$) and generate all $8$ targets for each identity.

\noindent\textbf{Downstream Details.}
\begin{itemize}
    \item \textit{Controllable face generation.}   For all experiments, we freeze the face representation encoder and optimize both the ControlNet and IP-Adapter using a constant learning rate of $2e^{-5}$ and an effective batch size of $16$ for $15$K steps. The models are trained on MFHQ with a variable number of inputs (uniformly sampled between $1$ and $7$) and a single target image, all at a resolution of $512\times 512$. All models converge well before reaching $15$K steps. The ControlNet is implemented and initialized as described in \cite{flux_controlnet}. The IP-Adapter is initialized from our Flow Matching Decoder. For fair comparisons, other representations (e.g., CLIP and ArcFace) also undergo the Flow Matching Decoder training stage to achieve convergence, requiring $5$K steps for CLIP and $75$K steps for ArcFace. In the benchmark, ground truth landmarks from the same identity are used as ControlNet inputs, and metrics are calculated between the generated images and the targets. The generation resolution is set to $512\times 512$. \\
    
    \item \textit{Personalized T2I generation.} 
    We integrate frozen face representations into the frozen FLUX dev base model~\cite{FLUX} using learnable decoupled attentions, following the approach outlined in IP-Adapter~\cite{ye2023ip-adapter}. Injecting into MM-DiT blocks is unnecessary in personalized T2I and does not affect the image quality. The IP-Adapter is trained using a simple flow-matching loss without additional regularization (\eg ID loss, alignment loss \cite{PuLID}) and without employing LoRA~\cite{hu2022lora}. These regularization and LoRA modules are left for future study as orthogonal to our work. Our training is performed at a resolution of $512\times 512$ for 50K steps with a constant learning rate of $1e^{-4}$, using the AdamW optimizer. Subsequently, we fine-tune the IP-Adapter at a resolution of $768\times 768$ for 20K steps, maintaining the same hyperparameters. Models are trained on our internal purchased dataset (Getty Images). For fair comparisons, other representations, such as CLIP and ArcFace, are trained under the same hyperparameters unless otherwise noted. Due to its slower convergence compared to Omni-ID and CLIP, ArcFace requires 100K steps in the first stage to achieve convergence. Inference for this task is performed at a resolution of $1024\times 1024$.
\end{itemize}

\begin{figure}[t] 
\centering
\includegraphics[width=0.91\linewidth]{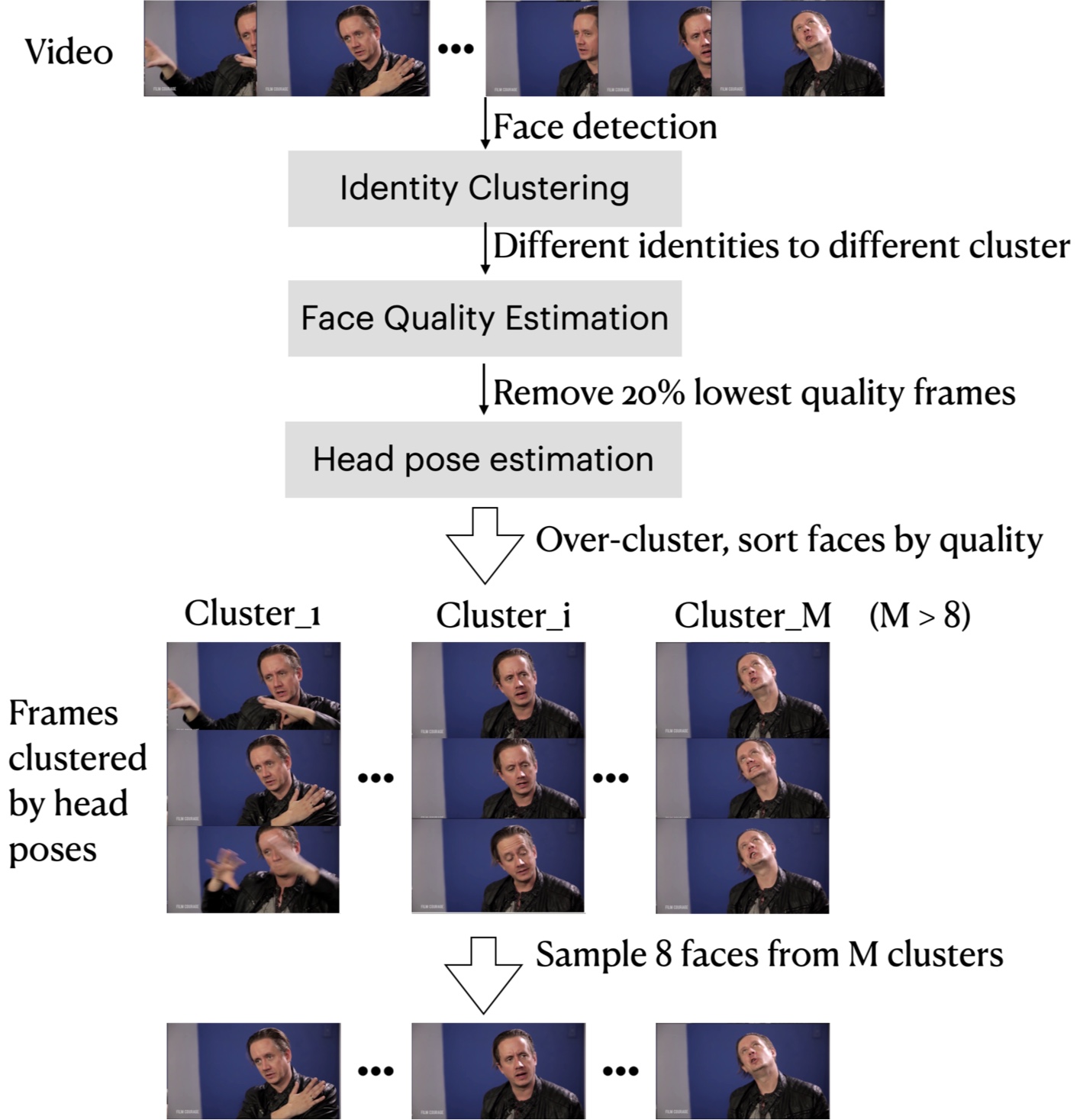}
\caption{\textbf{Illustration of MFHQ Creation}. Given a video, we first detect faces and distribute identities to different clips by a threshold based on the cosine distance of face embeddings \cite{ArcFace}. Then, a face quality estimation \cite{chen2024topiq} is applied to sort the quality of frames within each identity. 20\% faces with lowest quality are removed. A head pose estimation \cite{HopeNet} is employed to estimate the poses for each face which are used to cluster the frames into $M=16$ clusters. Finally, $8$ frames are sampled $M$ clusters, where each cluster is only sampled at most once. The sum of absolute pose differences is assured larger than $15$ degree for each pair.  
}
\label{fig:supp:mfhq_create}
\end{figure}

\noindent\textbf{MFHQ Details.} Refer to  \cref{fig:supp:mfhq_create} how  MFHQ is collected for each video clip. 

\section{Supplementary Experiments}

\subsection{Additional Controllable Face Generation}

\begin{figure*}[t] 
    \centering \includegraphics[width=0.95\textwidth]{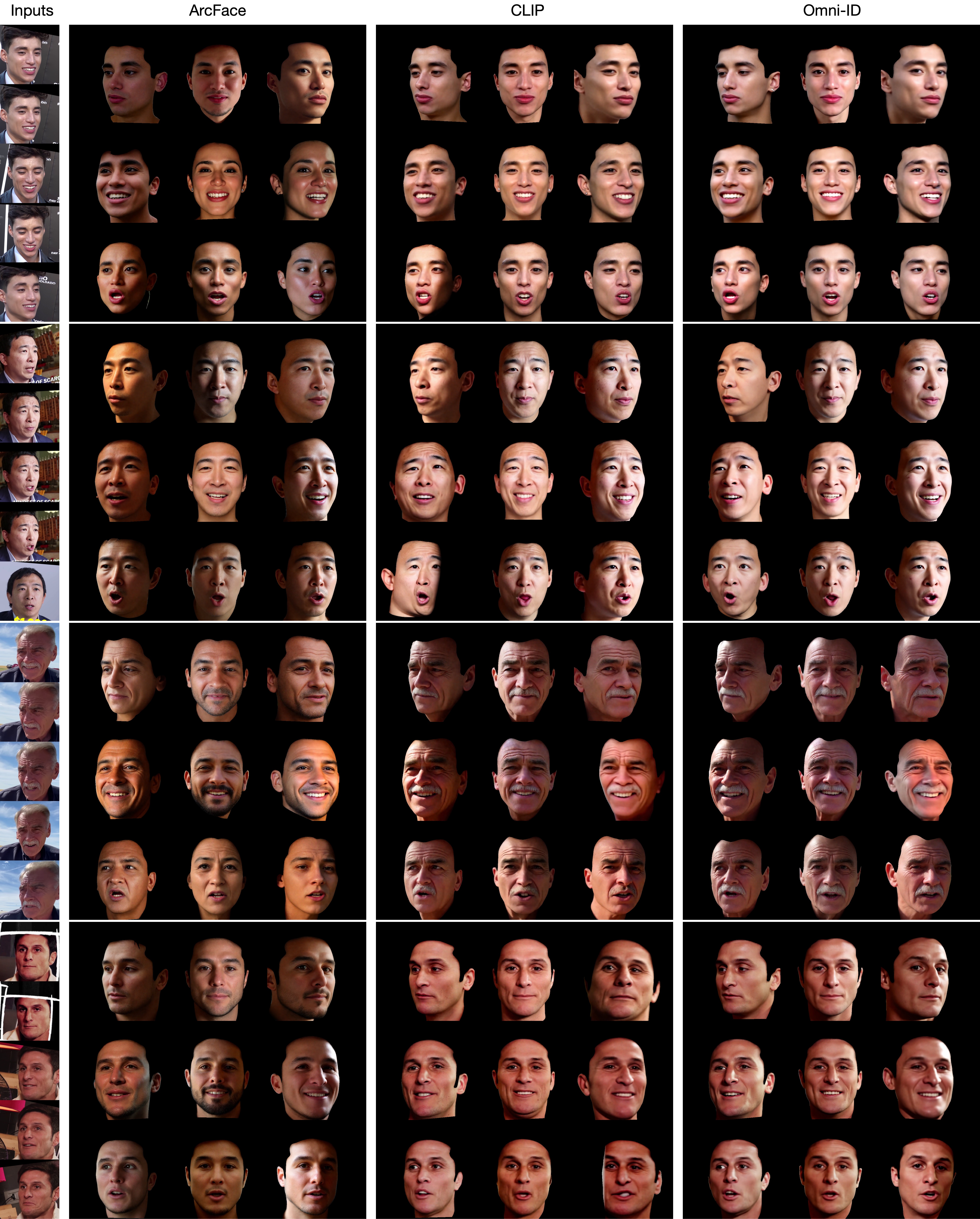}
    \caption{\textbf{Qualitative comparisons to the state-of-the-art representations in controllable face generation.} 
    We compare \methodname~with ArcFace~\cite{ArcFace} and CLIP~\cite{CLIP} with 5 input images. To control each face in the grid, we drive the facial landmark of each identity by the same template. Our \methodname~achieves superior identity preservation, captures nuanced details more faithfully, and demonstrates higher adaptivity to diverse poses and expressions. 
    }
    \label{fig:supp:face_grid}
\end{figure*}

\cref{fig:supp:face_grid} further compares \methodname~with ArcFace~\cite{ArcFace} and CLIP~\cite{CLIP} in the context of controllable face generation. Unlike the benchmark case presented in the main paper, where Ground Truth landmarks were used to guide identity-specific generation, here we use the template-driven landmarks as conditions. 9 template images are collected to obtain a grid of expression and pose in FLAME code \cite{FLAME:SiggraphAsia2017} through 3D mesh reconstruction by 3D landmark estimation. Then, we use the FLAME shape code for each identity with the template FLAME expression and pose code from each template to get the rigged mesh. From each mesh, 2D lanmarks are rendered as condition to generate each view at the grid for each identity.

While CLIP demonstrates strong baseline performance, it struggles with identity preservation and fails to generate realistic faces when the pose and expression differ significantly from the input images. This limitation arises because CLIP is an instance-level representation model. In contrast, our Omni-ID is an identity-level representation, specifically trained to reconstruct faces in new poses and expressions. Consequently, Omni-ID achieves significantly better identity preservation while generating new faces of the identity.

\subsection{Additional Personalized Text-to-Image}

\begin{table}[t]
\centering
\caption{\textbf{Quantitative comparisons to the state-of-the-art on personalized T2I generation.} ID Similarity are computed by the cosine distance between the generated samples and the five images of each identity. We compute the average and std across identities. The base models are FLUX~\cite{FLUX} for all methods. 
}
\label{tab:supp:personlization}
\begin{tabular}{@{}lc@{}}
\toprule
\textbf{Method} & \textbf{ID Similarity}$\uparrow$ \\ 
\midrule

IPA-FaceID & 0.3535$\pm$0.1982 \\

IPA-FaceIDPlusV2 & 0.4327$\pm$0.1090 \\ 

IPA-Full &  0.6649$\pm$0.0846 \\

PuLID & 0.7289$\pm$0.0572 \\

\textbf{IPA-Omni-ID Schnell (Ours)} & 0.7306$\pm$0.0793 \\

\textbf{IPA-Omni-ID (Ours)} & \textbf{0.8026}$\pm$\textbf{0.0421} \\
\bottomrule
\end{tabular}%
\end{table}

\begin{figure*}[t] 
\centering
\includegraphics[width=0.98\linewidth]{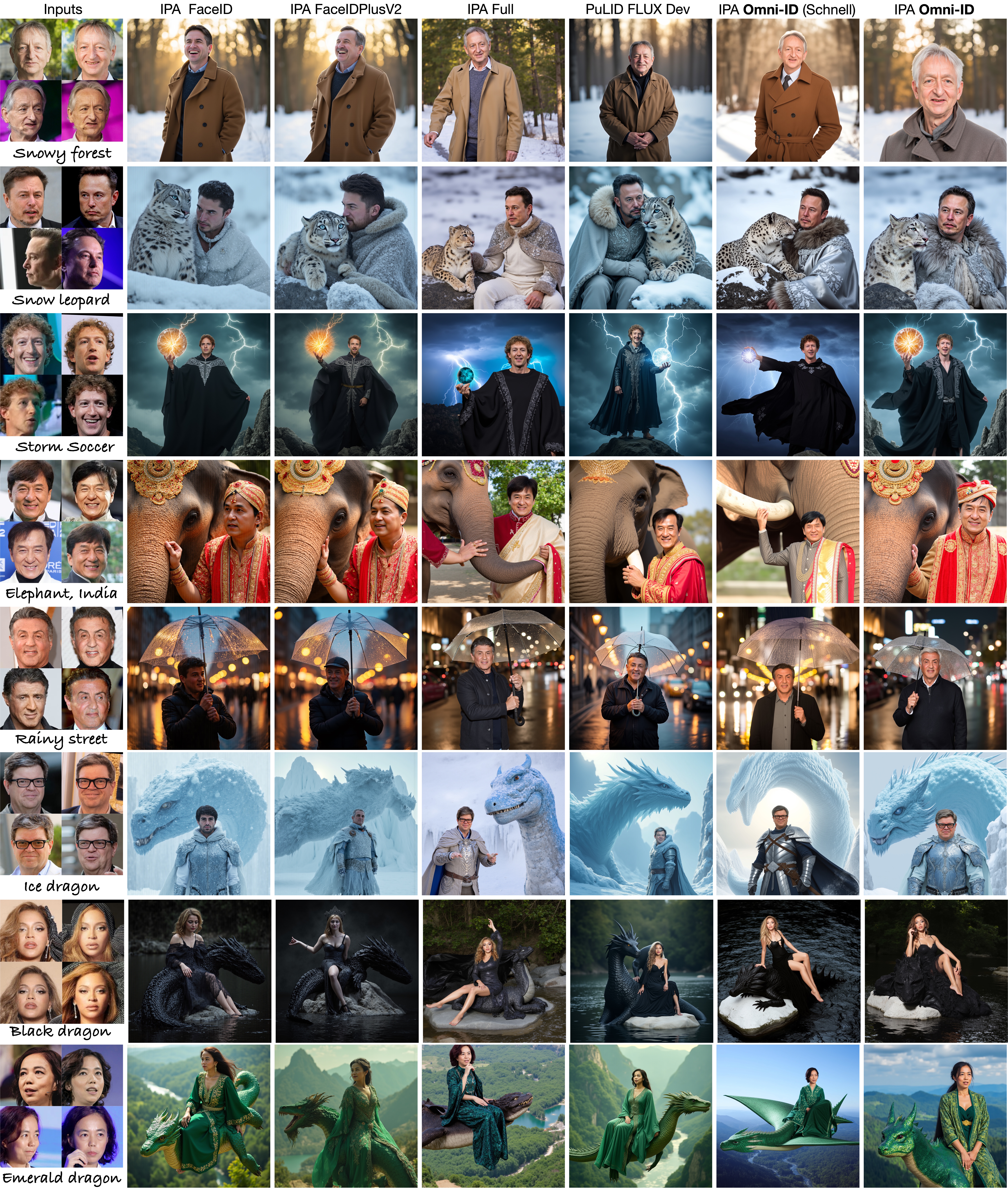}
\caption{\textbf{Qualitative comparisons with the state-of-the-art in personalized T2I generation using FLUX \cite{FLUX} as the base model.} Our \methodname~with IP-Adapter~\cite{ye2023ip-adapter} without any other regularization (LoRA \cite{hu2022lora}, ID loss \cite{PuLID}, alignment loss \cite{PuLID}) achieves highest ID preservation. 
Different variants of IP-Adapter without LoRA are shown at the left side. The state-of-the-art PuLID-FLUX-v0.9.1 achieves lower face quality compared to Omni-ID. Omni-ID also works well on FLUX Schnell model, which generates each sample by $4$ denoising steps.
}
\label{fig:supp-sota-person-flux}
\end{figure*}

\begin{figure*}[t] 
\centering
\includegraphics[width=1.0\linewidth]{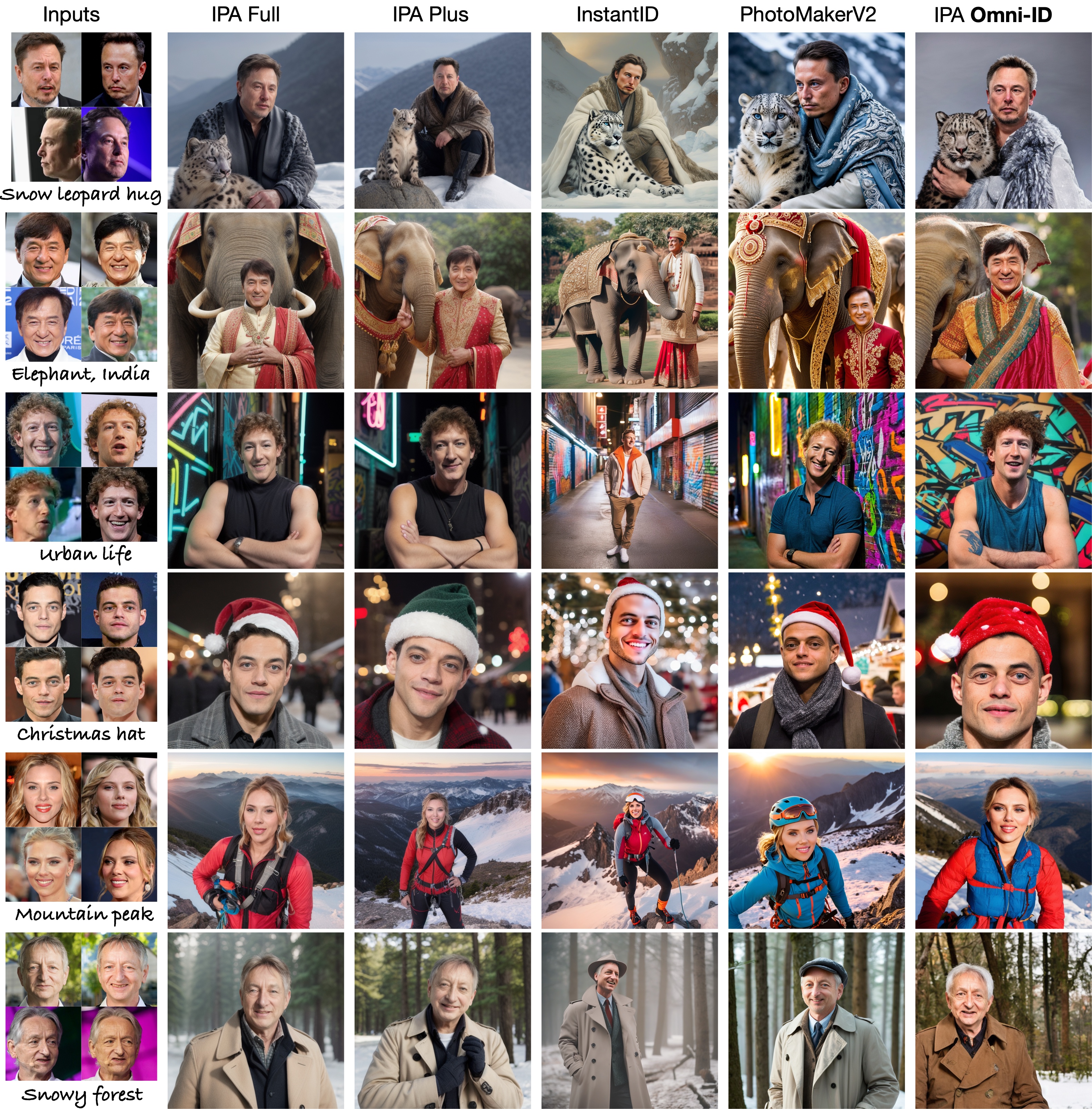}
\caption{\textbf{Qualitative comparisons to the state-of-the-art in personalized T2I generation using Stable Diffusion \cite{ldm} as the base model.} IPA-Full, IPA-Plus, and our IPA-Omni-ID use SD15 \cite{ldm} as the base model, generating $512 \times 512$ resolution samples. InstantID \cite{wang2024instantid} and PhotoMakerV2 \cite{li2023photomaker} use SDXL \cite{SDXL} as the base model, generating $1024 \times 1024$ samples, which are resized to $512 \times 512$ to show with other methods side by side. Our \methodname~with IP-Adapter without any other regularization achieves the highest ID preservation. 
}
\label{fig:supp-sota-person-sd}
\end{figure*}

\noindent\textbf{Compare to State-of-the-Art.} 
We compare Omni-ID+IP-Adapter (IPA Omni-ID) to the state-of-the-art IP-Adapter~\cite{ye2023ip-adapter}, InstantID~\cite{wang2024instantid}, PhotoMakerV2~\cite{li2023photomaker}, PuLID~\cite{PuLID} in \cref{fig:supp-sota-person-flux} and \cref{fig:supp-sota-person-sd} when using FLUX Dev \cite{flux_controlnet} and Stable Diffusion (SD)~\cite{ldm} as the base model, respectively. 
Our IPA Omni-ID trained by the simple flow matching loss without any advanced techniques such as LoRA \cite{hu2022lora}, ID loss \cite{PuLID}, aligment loss \cite{PuLID}, stacked embedding \cite{li2023photomaker}, IdentityNet \cite{wang2024instantid}, achieves the highest ID preservation. Refer to \textit{gallery.m4v} for all visual results of our model. \cref{tab:supp:personlization} compares IPA Omni-ID with the state-of-the-art personalized T2I employed FLUX as the base model. Our IPA Omni-ID outperforms others with the highest identity similarity.









\noindent\textbf{Beyond FLUX Dev Experiments.}
Despite Omni-ID is trained using FLUX dev~\cite{FLUX} as the Flow Matching Decoder, Omni-ID can be applied to any other diffusion models. In this section, we use the Omni-ID encoder with IP-Adapter on FLUX Schnell~\cite{FLUX} and SD15~\cite{ldm} in the task of personalized text-to-image generation. \cref{fig:supp-sota-person-flux} and \cref{fig:supp-sota-person-sd} demonstrates again the superiority of Omni-ID against other representations like CLIP and ArcFace.

\end{document}